# Around the GLOBE: Numerical Aggregation Question-Answering on Heterogeneous Genealogical Knowledge Graphs with Deep Neural Networks

Omri, OS, Suissa

Bar Ilan University, Department of Information Science, Ramat Gan 52900, Israel, omrishsu@gmail.com

Maayan, MZG, Zhitomirsky-Geffet

Bar Ilan University, Department of Information Science, Ramat Gan 52900, Israel, maayan.zhitomirsky-geffet@biu.ac.il

Avshalom, AE, Elmalech

Bar Ilan University, Department of Information Science, Ramat Gan 52900, Israel, avshalom.elmalech@biu.ac.il

One of the key AI tools for textual corpora exploration is natural language question-answering (QA). Unlike keyword-based search engines, QA algorithms receive and process natural language questions and produce precise answers to these questions, rather than long lists of documents that need to be manually scanned by the users. State-of-the-art QA algorithms based on DNNs were successfully employed in various domains. However, QA in the genealogical domain is still underexplored, while researchers in this field (and other fields in humanities and social sciences) can highly benefit from the ability to ask questions in natural language, receive concrete answers and gain insights hidden within large corpora. While some research has been recently conducted for factual QA in the genealogical domain, to the best of our knowledge, there is no previous research on the more challenging task of numerical aggregation QA (i.e., answering questions combining aggregation functions, e.g., count, average, max). Numerical aggregation QA is critical for distant reading and analysis for researchers (and the general public) interested in investigating cultural heritage domains. Therefore, in this study, we present a new end-to-end methodology for numerical aggregation QA for genealogical trees that includes: 1) an automatic method for training dataset generation; 2) a transformer-based table selection method, and 3) an optimized transformer-based numerical aggregation QA model. The findings indicate that the proposed architecture, GLOBE, outperforms the state-of-the-art models and pipelines by achieving 87% accuracy for this task compared to only 21% by current state-of-the-art models. This study may have practical implications for genealogical information centers and museums, making genealogical data research easy and scalable for experts as well as the general public.

CCS CONCEPTS • Computer systems organization~Architectures~Other architectures~Neural networks • Information systems~Information retrieval~Retrieval tasks and goals~Question answering • Computing methodologies~Artificial intelligence~Natural language processing

**Additional Keywords and Phrases:** numerical aggregation question answering, transformers, deep neural networks, genealogical domain, data modeling, knowledge graph, GEDCOM, cultural heritage research, digital humanities.

## 1. INTRODUCTION

In the past two decades, there has been an increasing interest in the construction and investigation of genealogical databases. For example, commercial companies like My Heritage and Ancestry collect over 48 million[i] and 100 million[ii] genealogical family trees, respectively; FamilySearch hosts over a billion[iii] unique individuals in the most significant non-profit collection of family trees. Family trees can be generated using different data sources, such as user-generated content ("personal heritage") [6], biographical registers [45], DNA records and clinical reports [13, 63], and even harvested from books [22; 96]. Online search services built upon genealogical data provide users with rich information about individual members and their genealogy relationships. In addition, these databases can be useful for population and migration research [55], historical preservation [33], and even for medical usage [79, 80].

Natural language search is a widespread practice that enables scholars (and the general public) to explore cultural heritage corpora [26]. However, in the genealogy domain (and other similar domains), database investigation based on search has some well-known limitations, as users must decompose their questions into keywords and then manually scan the obtained results to retrieve the specific information of their interest [26]. Moreover, when distant reading is required, a researcher must collect data from a long list of search results and perform the required calculations manually. Therefore, QA algorithms have been developed to enhance search systems. These algorithms receive questions in natural (human) language and return precise answers to these questions. Deep neural networks (DNNs) trained on large datasets are currently the state-of-the-art method for the QA task. These DNN-based QA systems allow humanities and social sciences researchers (and the general public) with no mathematical or programming background to ask research questions on cultural heritage data in natural language and receive precise answers to these questions.

There are seven types of questions mentioned in previous research: 1) Factual questions (what, when, which, who, how) that refer to a single answer (e.g., Who was the first person to classify birds?), 2) Numerical reasoning / numerical aggregative / arithmetic questions are factual questions that require a numerical calculation (e.g., What is the average number of birds traveling from Mexico to the United States every summer?), 3) List questions are factual questions that refer to a list of answers (e.g., Which birds types have blue wings?), 4) Definition questions that refer to a summarization of a topic (e.g., What is a bird?), 5) Hypothetical questions that require information associated with any assumed event (e.g., What would happen if birds had feet?), 6) Causal questions (how or why) that seek an explanation, reason, or elaboration for specific events or objects (e.g., Why birds fly?), 7) Confirmation questions that seek a confirmation (yes/no) of a specific fact (e.g., Can birds fly?) [54, 35, 12, 48, 41, 50, 21].

From an automatic QA perspective, factual natural questions constitute the most researched type of questions that have been extensively studied in the literature with high-accuracy results [1, 11, 60, 73, 77, 78, 81]. Factual questions concentrate on a specific item [34]. For instance, given a question (related to a particular person): "Where was John Doe's father born?", the answer is: "Kenya" (a specific attribute of a person). Training DNN models for factual QA is done using a dataset that comprises triples of the form: (question, answer, corresponding text passage), from which the answer can be extracted.

Unlike factual natural questions, numerical aggregation questions, the focus of this paper, pertain to a group of items in the dataset and require applying mathematical computation (an aggregation function) to the matching items. For example, for the question "What is the average age of men in John Doe's family that were born from 1790 to 1860?", a numerical aggregation QA system should return the answer "61.5" by calculating the average value of the "age" attribute of all the men born between 1790 and 1860. Answering these types of questions is critical for conducting distant reading analysis of corpora and allows researchers to answer questions that otherwise require too much human effort and time.

Training DNN models for numerical aggregation QA requires a golden standard dataset that comprises questions, answers, and table/s of data from which these answers can be computed, where the answer is the result of an aggregation function (e.g., average, sum, count, min, max) on some of the table's rows. Such a table-based approach can also be applied to factual QA [27, 31, 32, 82, 38]. However, aggregation natural QA is considered a more challenging task, since DNN models for answering numerical aggregation questions do not only have to select the relevant cells from the table/s (as factual QA models do), but also need to learn and compute the aggregation functions that are implied from the question (i.e., perform numerical reasoning). For example, for the question "How many people in my family lived in England?", the model needs to *count* the number of rows matching a criteria (e.g. "lived in England"), and for the question "What is the life expectancy of men in my family tree?", the model needs to calculate the *average* value of a specific cell (i.e., age) in the rows matching a criteria (e.g. "gender is male").

The genealogical domain poses several additional challenges for QA DNNs, as there are no available training datasets, and some adaptations of the existing models are needed since genealogical data is usually stored as a GEDCOM (GEnealogical Data COMmunication) graph, rather than a set of texts or tables as expected by the existing QA models; the data is comprised of numerous linked entities (data on persons and families and their multi-level inter-relationships) which overall size may exceed standard DNN's input limitation; the questions are mostly on the relationships rather than just about entities (i.e., persons) in the graph, and require additional aggregation function types that have not been implemented in the existing models.

Hence, the main objective of this paper is to design and evaluate a novel methodology for the DNN-based numerical aggregation QA task in the genealogical domain. Cultural heritage corpora are highly suitable for using different natural language processing (NLP) algorithms to support research. The vast amount of texts, images, and other data types in these corpora can be analyzed and used to extract insights and answer many research questions [68]. For example, [10] researched visual QA on cultural heritage corpora when a user provides an image and asks a question on that image [10]; [65] researched a QA chatbot for answering cultural heritage questions for tourists; and [7] used question generation and answering to create a "self-managed" corpus. While there is research on QA in various cultural heritage domains, including genealogy [69], to the best of our knowledge, this is the first research on numerical aggregation QA (that is critical for quantitative analysis and distant reading of corpora) for cultural heritage and specifically for the genealogical domain. The developed methodology is referred to as GLOBE (Genealogical Legacy Overview with Bert Embeddings) and comprises the following main components: 1) a new automated method for dataset(s) generation for numerical aggregation QA based on the knowledge graph representation of genealogical data, 2) a fine-tuned DNN method for optimal table selection based on SBERT [58]; and 3) a fine-tuned numerical aggregation QA DNN model for the genealogical domain, based on BERT [16]. We experimented with six different tabular data models and evaluated the influence of the dataset structure and quality on the DNN's accuracy. The proposed methodology increases the amount and complexity of genealogical data that the QA DNN model can use and outperforms the state-of-the-art aggregation QA model when applied to the genealogical datasets, thus showing the benefit of the domain-specific approach for the task [69].

## 2. RELATED WORK
This section reviews relevant work in the fields of genealogical data representation, DNN architecture, and numerical aggregation QA using DNN.

### 2.1. Genealogical data representation
Developed in 1984 by The Church of Jesus Christ of Latter-day Saints, the GEDCOM format is the de facto standard for data representation in the genealogical field [24, 30, 39]. The GEDCOM format has a simple linkage-based structure where records containing names, events, places, relationships, and dates are arranged hierarchically [24].

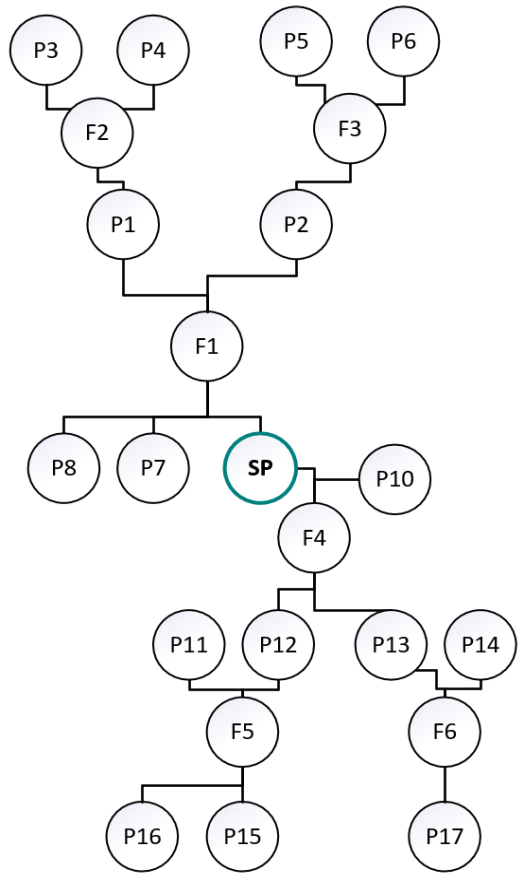

Figure 1: Family tree structure [69].

```
0 HEAD
1 SOUR SomeSite
2 NAME Some Site
2 VERS 3.0
1 DATE 28 JUN 1984
2 TIME 16:18:10
1 FILE example2.ged
1 GEDC
2 VERS 5.5
2 FORM LINEAGE-LINKED
...
1 CHAR ANSEL
0 @I138@ INDI
1 NAME Mary Lulu
1 SEX F
1 BIRT
2 DATE 27 MAY 1756
2 PLAC New Jersey, USA
1 DEAT
2 DATE 7 FEB 1815
2 PLAC Philadelphia, USA
1 BURI
2 DATE 8 FEB 1815
2 PLAC Philadelphia, USA
1 BAPL
2 DATE 1 JUN 1776
1 ENDL
2 DATE 30 DEC 1775
2 Christ Church
```

Figure 2: A fragment of the GEDCOM family tree file displayed in Figure 1.

In GEDCOM, every individual (person) in the family tree is represented as a node containing predefined attributes, such as name, birth date and place, death date and place, burial date and place, occupation, and other details. Every individual is a "spouse" (i.e., a parent) or a "child" within a family node. Figure 1 shows a sub-graph corresponding to a Source Person (SP) whose data is displayed in the GEDCOM file in Figure 2. A number bracketed between @ symbols and a class name (INDI – individual, FAM – family) is assigned to every person and family node. The source person is denoted as SP (e.g., @I138@ INDI Mary Lulu in the GEDCOM file), families as F, and other persons as P. As shown in Figure 2, @I138@ (i.e., Mary Lulu) was a female, born on 27 MAY 1756 in New Jersey, USA, who died on 7 FEB 1815 in Philadelphia, USA, and was buried a day later in the same place.

**2.2. Numerical Aggregation QA using DNNs**

DNN models have become a standard method for developing natural language QA systems in recent years [43]. Numerical aggregation QA is a relatively new, challenging, and underexplored task. There are two main approaches for this task: answering the question directly using a DNN model [27, 32, 82, 38, 31] or converting the question to a formal language query (e.g., in SQL or SPARQL) and using a formal language parsing engine to calculate the answer (i.e., executing the query) [2, 17, 28, 87, 76]. The former approach has a considerable advantage: answering the question is an approximation task with a straightforward output (i.e., a number), while the feasibility and applicability of the latter approach are still under discussion [59] as formal language queries are a much more complicated output. For example, for the question "what is the portion of women in my family tree that were single over the age of 30?" the first approach produces the output of: *"6.43%"*, while the second approach returns the SQL query "*SELECT (COUNT(\*) / (SELECT COUNT(\*) FROM table_person WHERE gender = N'F')) as 'Portion' FROM table_person WHERE gender = N'F' AND (marriage_year - birth_year) > 30"*. Moreover, converting a natural language question to a formal language query requires a massive amount of training data (i.e., pairs of natural language questions and corresponding formal queries).

Numerical aggregation QA models have achieved varying accuracy in multiple studies based on paragraphs [4, 25, 57] and structured tables represented as text [32, 5]. Several open-domain datasets have been generated to evaluate table-based QA DNN models (for both factual and numerical aggregation questions). For instance, WikiTQ [56] comprises complex questions from Wikipedia tables; SQA [36] is a conversational dataset created by crowdsourcing from the WikiTQ dataset; and WikiSQL [87] is a formal query paraphrasing mapped to a natural text dataset. The models' accuracy depends on the complexity of the dataset and question type (higher accuracy is achieved for factual questions than for numerical aggregation questions), and varies between 33% and 86% (Table 1).

Table 1: Accuracy reported for the state-of-the-art QA models on open-domain datasets.

| Model | WikiSQL | WikiTQ | SQA (Q1) | SQA (AVG) |
| --- | --- | --- | --- | --- |
| Pasupat and Liang, 2015 | - | 37.1 | 51.4 | 33.2 |
| Iyyer et al., 2017 | - | - | 70.4 | 44.7 |
| Neelakantan et al., 2017 | - | 34.2 | 60.0 | 40.2 |
| Zhang et al., 2017 | - | 43.7 | - | - |
| Liang et al., 2018 | 71.8 | 43.1 | - | - |
| Haug et al., 2018 | - | 34.8 | - | - |
| Liang et al., 2018 | - | 43.1 | - | - |
| Sun et al., 2018 | - | - | 70.3 | 45.6 |
| Agarwal et al., 2019 | 74.9 | 44.1 | - | - |
| Wang et al., 2019 | 79.4 | 44.5 | - | - |
| Min et al., 2019 | 84.4 | - | - | - |
| Dasigi et al., 2019 | - | 43.9 | - | - |
| Muller et al., 2019 | - | - | 67.2 | 55.1 |
| Herzig et al., 2020 | 85.1 | 42.6 | **78.2** | 67.2 |
| Bagwe et al., 2020 | - | 44.5 | - | - |
| Krichene et al., 2021 | **85.82** | 49.13 | - | - |
| Eisenschlos et al., 2021 | - | **51.5** | - | **71.7** |

Since DNNs perform slowly in predicting answer spans (in a factual QA task) or selecting cells (in a numerical aggregation QA task) for a given passage of text or table, they are not applied to the entire database, but only to selected pieces of data [1]. Moreover, while some DNN models can accept a large input [8, 40], many state-of-the-art DNN models (such as BERT) tend to accept a limited size input, usually ranging from 128 to 512 tokens (i.e., words) [23] due to computational resource limitations.

Therefore, there is a need to develop an optimal data selection method before applying the DNN model. Thus, in the factual QA task, given a user's question, the system retrieves the top K passages in the dataset that are relevant to the question (typically based on a reverse indexing approach [9, 37, 49, 62]). Using the DNN model, the system then predicts the answer spans (start and end positions) for each of the K-selected passages with a certain confidence level.

However, when answering a numerical aggregation question, the model must receive all the data to perform the numerical function calculation (e.g., count, average). Hence, earlier works suggested splitting the input into cells, rows or columns to overcome this limitation. [5] proposed classifying rows based on the given question and passing only selected rows to the model. [18] devised a heuristic-based column selection technique. [42] designed a model-based cell selection technique that is differentiable and trained with the main task model. [85] used an attention mask to restrict attention to tokens in the same row and column, and [19] suggested attention heads to reorder tokens based on a row or column. However, these methods limit the aggregation functions that can be computed, as they exclude the cases when the aggregation function is relative to the entire table (e.g., average over the whole table). Alternatively, [31] used a DNN model to infer an answer for each table in the dataset (or top K tables), and the answer with the highest confidence score was selected.

In the genealogical domain, there is a relatively small number of entities (i.e., node types, such as a person or family) and a large number of relationships, which, when converted into the tabular data format, may yield very large tables. In addition, aggregative QA in the genealogical data requires joining multiple tables (e.g., to answer the question "How many people become parents before the age of 20?" we need both the person's age and his first child's birthdate). Therefore, the methods mentioned above that perform a single table selection are not applicable in the genealogical domain that requires joining several tables to support aggregation QA. As a result, the generic approaches described above cannot be applied as-is in the genealogical domain without reducing accuracy, but require certain modifications and adaptations. In addition, various training methods, hyperparameters, and architectures indicate the complexity of the task and its sensitivity to a specific domain [70].

## 3. METHODOLOGY

The proposed end-to-end pipeline of GLOBE consists of two parallel processes: the interactive inference process (Figure 3) and the model training, which runs in the background (Figure 4). As shown in Figure 3, the interactive inference process, using a simple user interface, allows users to select a family tree (1) and ask a question (2). Then, the system selects the table/s from the dataset that include/s the answer to the posed question (3). At this stage, in order to suit the amount of input data for the DNN's size limit, the user may be required to further reduce the scope of the explored family tree (i.e., a user interface that allows the user to select a relation degree[iv] is presented and the user selects a smaller degree) (4). Finally, if the input size is suitable, the system calculates the answer and displays it to the user (5).

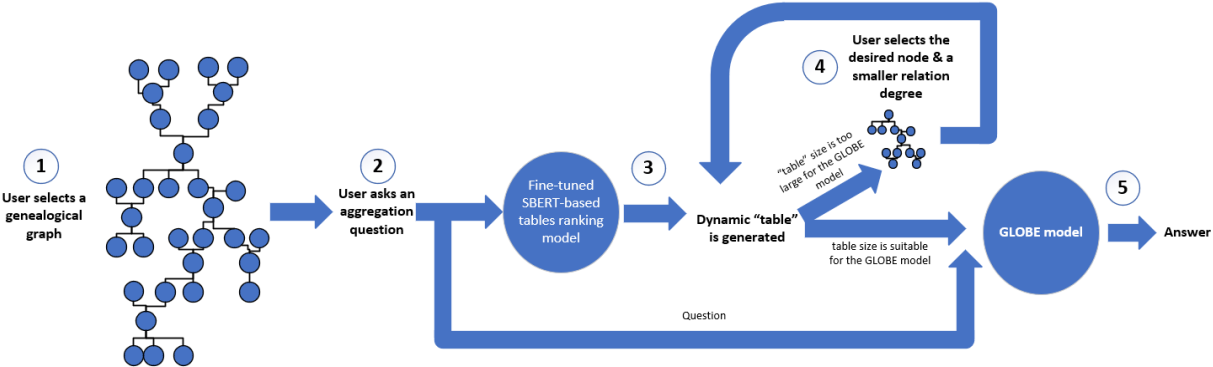

Figure 3: The GLOBE inference pipeline.

As shown in Figure 4, the proposed methodology for the model training comprises three main stages: 1) The generation of a table-based training dataset comprised of numerical aggregation questions from genealogical graphs, tables that contain data for answers to these questions and the corresponding answers; 2) Building a DNN table ranker for selecting the suitable table/s that contain/s the data for answering a given question; and 3) Building an optimal numerical aggregation QA DNN model using the best generated dataset and the table selection from the previous stages.

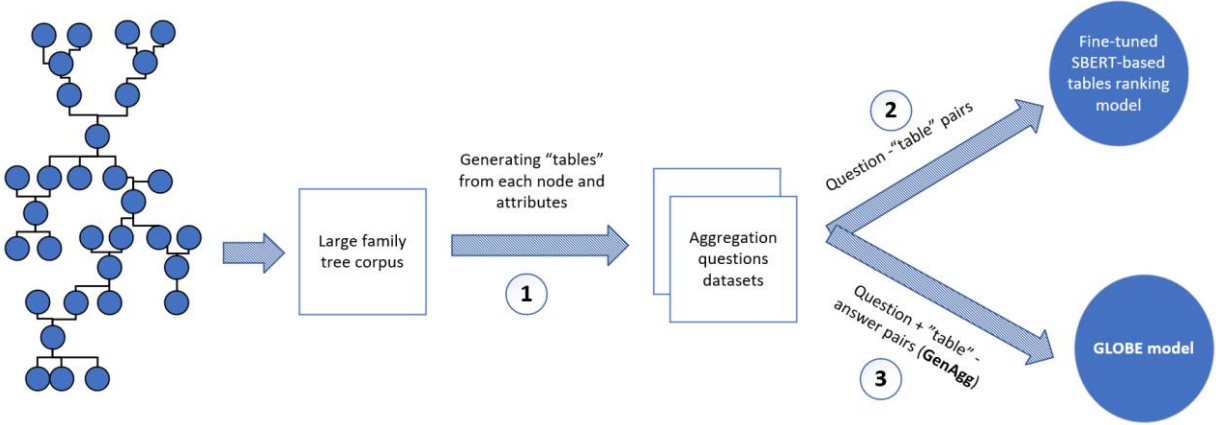

Figure 4: The GLOBE training pipeline.

Next, we describe each stage of the pipelines and the experimental setup of the study in more detail.

## 3.1. GenAgg Dataset Generation

Generating a training dataset from genealogical data is a two-step process. As shown in Figure 5, the method includes the following steps: (1) generating relational tables from the family tree knowledge graphs, and (2) generating questions and answers from the tables. The result of the process is a GenAgg dataset, a relational database composed of questions, tables with answers to these questions and answers tailored to the genealogical domain.

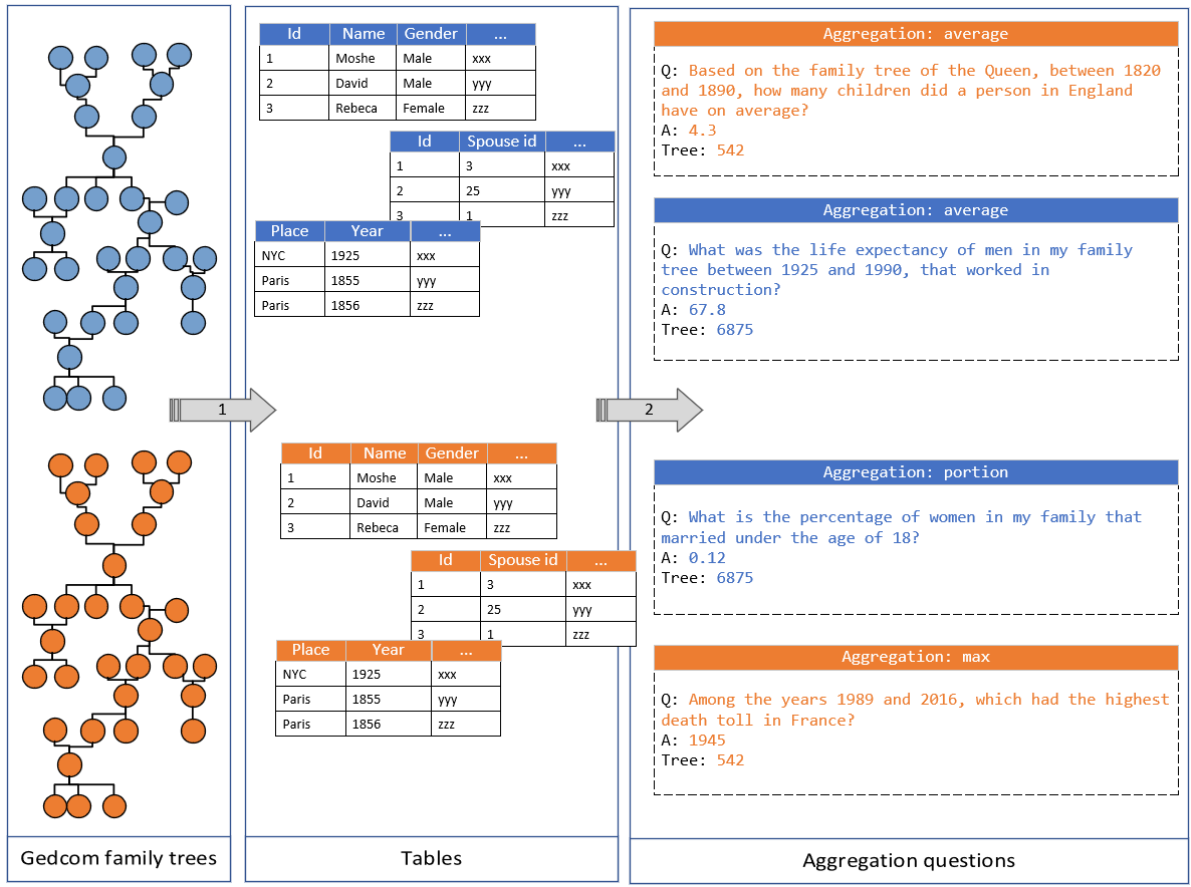

Figure 5: The GenAgg dataset generation process.

### 3.1.1. GEDCOM to Tables

GEDCOM graphs are first converted into CIDOC-CRM-based[v] formal knowledge graphs, as shown in [69]. These knowledge graphs are then represented as a relational database (GenAgg). While there are several alternatives for modeling a dataset (e.g., graph database [93, 94, 95], RDF triples, RDBMS), this paper uses a table-based approach (RDBMS) to be compatible with the state-of-the-art DNN model for aggregative question-answering (TaPas [32]). To determine the optimal database design (i.e., structure) for the task of numerical aggregation QA, we experimented with six alternative types of table structure (see Figures 6-11).

A single table structure (GenAgg$_{1t}$) is easy to implement and use, as many DNN models have been built to deal with this structure type [32, 38]. However, this design supports only persons with a single spouse, and requires the DNN model to infer some relationships (e.g., siblings, children) while answering questions.

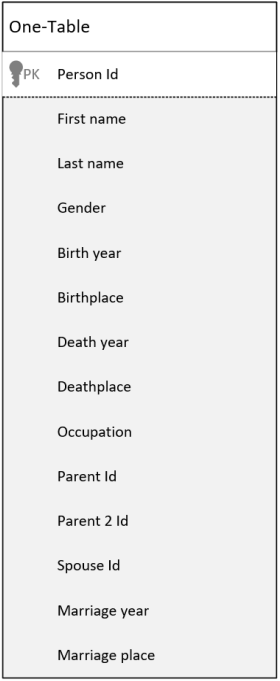

Figure 6: A single-table design of the database.

Figure 7 presents the "raw-data" design (GenAgg$_{raw}$). It contains two tables: the table of the persons (i.e., nodes in the graph and their attributes) and the table of kinship (i.e., pairs of persons and their relationships, edges in the graph). This design overcomes the limitation of spouses present in the single-table design; however, it lacks marriage attribute data and allows only Parent, Spouse, or Sibling relationships.

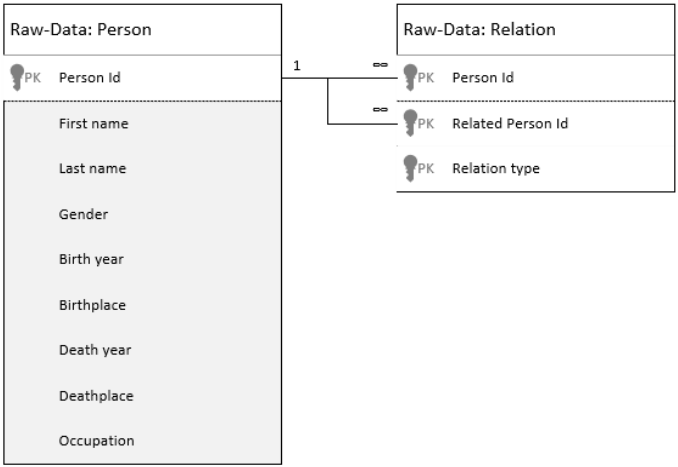

Figure 7: Raw-data design.

The third design, "relationship-driven" (GenAgg$_{rel}$), splits each relationship type (i.e., edge type) into a dedicated table with relevant attributes, thus reducing the model complexity when inferring relationships (Figure 8).

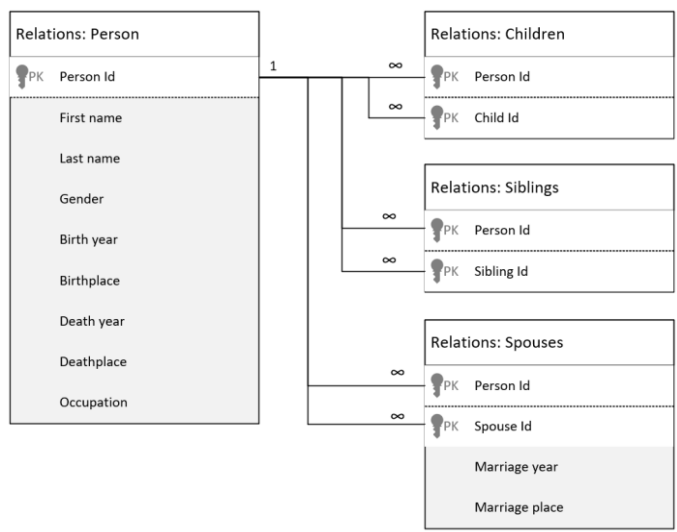

Figure 8: Relationship-driven design.

An unnormalized, pre-joined dataset can be created to enable more straightforward relationship inference. As shown in Figure 9, such an "aggregative" design (GenAgg$_{agg}$) duplicates each attribute and stores it in both persons' and relationships' tables. In addition, relationship counts (e.g., number of children) and the age of persons and relationships (e.g., when the first child was born) are pre-calculated and added to the tables to reduce the model's need to count each person's relationships.

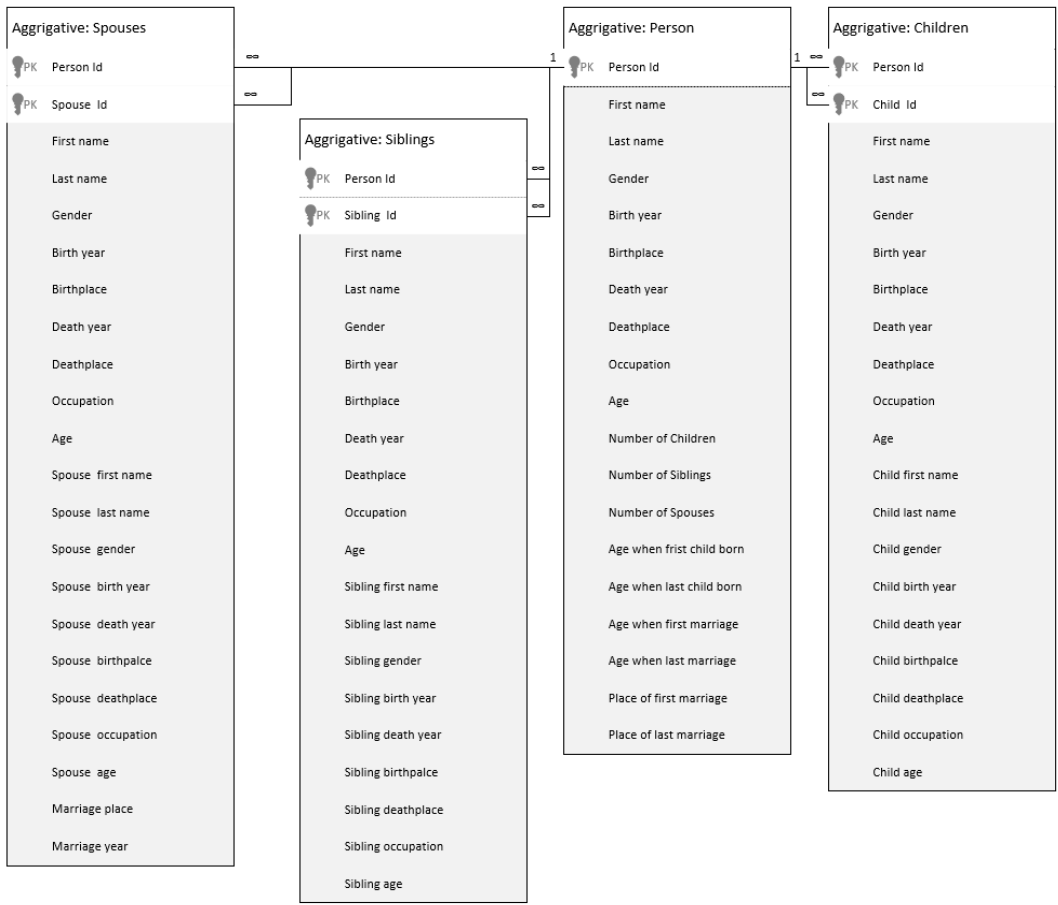

Figure 9: The "aggregative" design.

As shown in Figure 10, the "aggregative" design can be extended with an additional aggregation table for events (birth, death, and marriages) per year and place (GenAgg$_{event}$). These aggregations may simplify the inference for time and place-related questions.

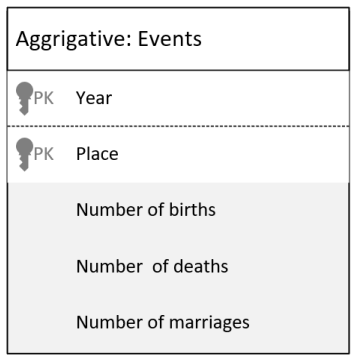

Figure 10: The aggregation table for events.

Finally, to further reduce the data size of the DNN input, a 6NF-based [15] (sixth degree of normalization - GenAgg$_{6NF}$) design was proposed where each non-primary key column in the "aggregative with events" design (Figures 9 and 10) will be moved into a separate table (Figure 11), thus creating a large number of tables with a small number (2-3) of columns.

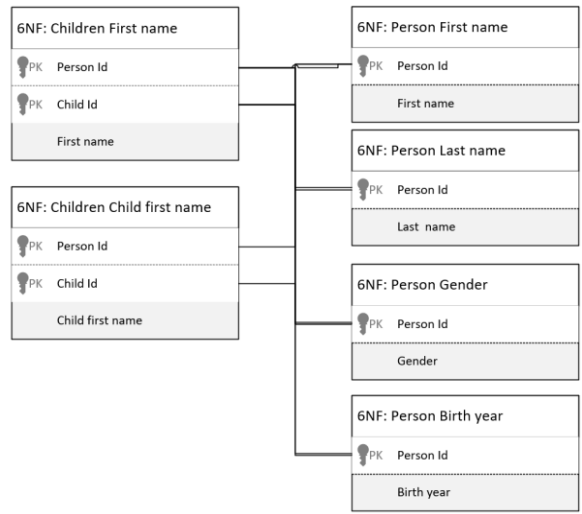

Figure 11: A segment of the "aggregative 6NF" design.

*3.1.2. Tables to Questions and Answers*

The next step in generating the training dataset is to compose questions and answers corresponding to the tables constructed above. This is done for each dataset design in two steps: first, the initial set of questions is compiled using a pattern-based approach, and then a DNN model is applied for rephrasing the questions to expand the initial question set. The patterns are manually created based on common research questions in the genealogical domain. To this end, each attribute has to be classified as a textual or numeric column that fits specific patterns. Each attribute can be used with certain predefined types of aggregation functions and conditions, and some attributes can serve as population descriptors (e.g., gender, occupation). For example, the column *age* is a numeric attribute; thus, it can be used with *min*, *max*, and *average* numerical aggregation functions, but it does not make sense with *sum*, and does not describe a type of population. Table 2 presents several examples of such patterns and the resulting questions. The initial set of questions can be systematically created with all the possible combinations of attributes and conditions using Algorithm 1. The algorithm iterates over the tables of the dataset (line 1), over the columns of each table (line 2), over pattern-formatting rules (as shown in Table 2) relevant to each column (line 3), and over the genders (i.e., male, female, both) (line 4). With the iteration's data, Algorithm 1 generates N conditions from other columns in the same table (lines 5-9); and sends the iteration's data and the conditions to the pattern-formatting rule (i.e., a function) that returns the question (line 10).

---

ALGORITHM 1: Pattern-based initial question generation

---

**Input:** Dataset ($D$), Table ($D_T$), Column ($D_{T_c}$), Max number of conditions ($N$)

**Output:** Question ($Q$)

**Initialization:** Pattern format rule ($Ps$), Genders ($Gs$)

1. foreach $D_T$ in $D$

2. foreach $D_{T_c}$ in $D_T$
3.   foreach $P$ in $Ps$
4.     foreach $G$ in $Gs$
5.       conds = []
6.       foreach $1..N$
7.         foreach $D_{T_c}2$ in $D_T$
8.           If $D_{T_c} != D_{T_c}2$
9.             conds.append($D_{T_c}2$)
10.       yield $Q = P(D_{T_c}, G, conds)$

Table 2: Pattern examples for question generation in the genealogical domain.

| Aggregation Function | Pattern-based rule example | Pattern input | Result |
|---|---|---|---|
| COUNT | How many [POPULATION DESCRIPTOR] were [CONDITION 1] … [and/or/between] [CONDITION N] | population descriptor: women<br><br>condition 1: born in England<br><br>condition 2: worked as a dressmaker<br><br>condition 3: between 1972 to 1982 | How many women were born in England and worked as a dressmaker between 1972 to 1982? |
| MIN | What is the minimum [COLUMN] for [POPULATION DESCRIPTOR] that were [CONDITION 1] … [and/or/between] [CONDITION N] | Column/attribute: age of marriage<br><br>population descriptor: women<br><br>condition 1: born in Germany<br><br>condition 2: between 1850 to 1900 | What is the minimum age of marriage for women that were born in Germany between 1850 to 1900? |
| MAX | What is the [COLUMN 1] with the maximum number of [COLUMN 2]s [were/in] [CONDITION 1] … [and/or/between] [CONDITION N] | Column 1: year<br><br>column 2: deaths<br><br>condition 2: in Germany | What is the year with the maximum number of deaths in Germany? |
| AVERAGE | What is the average [COLUMN] of [POPULATION | column: age<br><br>population descriptor: men | What is the average age of men that were born in Spain and worked as a |

|  | DESCRIPTOR] that were [CONDITION 1] … [and/or/between] [CONDITION N] | condition 1: born in Spain condition 2: worked as a pharmacist condition 3: between 1850 to 1950 | pharmacist between 1850 to 1950? |
| --- | --- | --- | --- |
| PORTION | What is the portion of [POPULATION DESCRIPTOR] that [CONDITION 1] … [and/or/between] [CONDITION N] | population descriptor: men condition 1: had three children | What is the portion of men that had three children? |

To avoid model overfitting and increase the language variability of the questions to better suit real-world applications, a question augmentation method based on a DNN model has to be employed [20]. To this end, the DNN model based on the PEGASUS [86] and trained on the Quora dataset [3] is utilized to paraphrase questions (see some examples in Table 3).

Table 3: Output examples of the question paraphrasing based on PEGASUS.

| Pattern-based rule input | PEGASUS paraphrasing output |
| --- | --- |
| How many women were born in England and worked as a dressmaker between 1972 to 1982? | Between the year 1972 and the year 1982, what was the number of women that held a dressmaker job and lived in England? |
| How many marriages were in Paris in 1950? | How many people got married in 1950 in Paris? |
| What is the minimum age of marriage for women that were born in Germany between 1850 to 1900? | For women from Germany, what is the min marriage age between the year 1850 and the year 1900? |
| How many men were born in Firenze, Italy? | How many men were born in Italy? |
| What is the year with the maximum number of deaths in Germany? | Which is the deadliest year in Germany? |
| What is the average age of men that were born in Spain and worked as a pharmacist between 1850 to 1950? | What is the life expectancy for pharmacist men from 1850 to 1950 in Spain? |
| What is the average number of births when the birth year was greater than 1985? | What is the average number of births when the birth year was less than 1985? |
| What is the portion of men that had three children? | What is the percentage of men with three children? |

Finally, the corresponding answer is extracted from the appropriate table using the question conditions and aggregation function. The resulting dataset is stored as a set of tuples of the form: (function, question, answer, table/s) as JSON files.

### 3.2. DNN-based Table Selection

In order to train the QA DNN model on the constructed datasets, table selection has to be performed, since the model can not get all dataset tables as input due to the BERT model input size limit (512 tokens). To this end, the SBERT model was fine-tuned on a subset of the above created datasets of questions and their source tables [58] to calculate the textual similarity between the given question and each of the tables' content in the dataset. SBERT seems suitable for this task since it was originally designed to calculate the similarity between a search query and its corresponding clicked document based on Siamese neural networks. The SBERT model was initially trained on the MS MARCO dataset that comprises 1,010,916 search queries and the corresponding clicked documents/passages [89]. To fine-tune the SBERT model for the genealogical domain, a dataset of 1,383,586 questions created as part of the GenAgg datasets was used; for each question, the source table(s) was saved as a positive (i.e., similar) example, and the other table(s) was saved as a negative (i.e., un-similar) example. Then, the top K (or less) similar tables are selected and joined using a set of join rules per dataset structure, and a dynamic, unified table is generated. If the new dynamic table size contains more than 512 tokens, the user can be asked to reduce the family tree by selecting the relational degrees in the scope using the Gen-BFS algorithm [69], or otherwise, the question can be returned as unanswerable. If the size of the resulting table is suitable for the model, it is set as an input for the QA DNN model.

### 3.3. DNN-based QA model for the genealogical domain

The proposed QA model is based on the weak supervision[vi] implementation of the state-of-the-art TaPas model [32], a BERT [16] encoder model adapted to the numerical aggregative QA over tables. TaPas has been designed to answer open-domain questions with three types of aggregation functions: count, sum, and average. The model uses a question and a flattened table as input; a table is flattened into a sequence of word pieces (tokens) and concatenated with the question tokens. The encoder model is added with two classification layers for selecting table cells and aggregation operators that operate on the cells. The cell selection classification layer determines whether (i.e., the probability of) a given cell should be used for the aggregative operation or not. The aggregation operator layer selects the numerical operation that is needed to answer the question. The input embeddings matrix comprises: (1) position id (like the BERT's index of the token in the flattened sequence), (2) segment id (0 for the question tokens and 1 for the table tokens), (3) column id (0 for the question tokens, column index for the table tokens), (4) row id (0 for the question tokens, row index for the table tokens), and (5) rank id (0 for non-numeric/date cells, order index for numeric/date cells). The embeddings improve the model's ability to understand the token's representation with respect to the question.

Table 4: The weak supervision implementation of GLOBE aggregation functions (f), where c is the cell scalar value in table T with selection probabilities ($p_s$) for the question and selection probabilities ($cp_s$) for the question population. Missing values are set to 0.

| GLOBE | TaPas | f | compute(f,$p_s$,$cp_s$,T) |
|---|---|---|---|
| ✓ | ✓ | COUNT | $\sum c \in T^{p_s^{(c)}}$ |
| ✓ | ✓ | SUM | $\sum c \in T^{p_s^{(c)}} \cdot T[c]$ |
| ✓ | ✓ | AVERAGE | $\frac{compute(SUM, p_s, T)}{compute(COUNT, p_s, T)}$ |
| ✓ | ✗ | PORTION | $\frac{compute(COUNT, p_s, T)}{compute(COUNT, cp_s, T)}$ |

As can be noticed from Table 4, GLOBE calculates all the functions implemented in TaPas. Note that min and max operations are considered cell selection operations and do not require any computation. In addition, GLOBE defines and implements a new function, *portion*, that was suggested by two interviewed genealogists as necessary for the genealogical domain. This function allows for computing the portion of a specified group compared to the entire population (e.g., "What is the portion of men married under 25 in the UK Queen's dynasty?"). Adding the portion function to the model's loss introduces a new challenge: determining the entire population relevant to the question. The portion function is calculated as the count of rows matching all the criteria in the question divided by the number of rows matching the entire relevant population. For example, the question above needs to be split into two sub-questions: 1) "What is the total number of men in the Queen's dynasty?" and 2) "What is the total number of men married under 25 in the Queen's dynasty?". The portion is the answer to the second question divided by the answer to the first question (i.e., $\frac{men\ married\ under\ 25}{men}$). Therefore, when calculating the loss, the first word in the question that matches the population type using a predefined list of population types is extracted, and the model is rerun in a non-trainable mode with a fixed pattern ("how many [POPULATION DESCRIPTOR]?"), forcing only the count function calculation. Then, the population cell probabilities are transferred back to the loss to calculate the portion. When no population values are found in the question or no rows are found in the second forward pass of the model, the portion is considered 0.

## 4. EXPERIMENTAL DESIGN

To implement and evaluate the effectiveness of the proposed methodology (i.e., finding the best-performing dataset design and model for answering genealogical numerical questions), a series of experiments were carried out as follows.

### 4.1. Datasets

The datasets in this research contained 1,847,200 different individuals from 3,139 genealogical trees (i.e., GEDCOM files) from the corpus of the Douglas E. Goldman Jewish Genealogy Center in the Anu Museum[vii]. The Anu Museum collection holds over 5 million different individuals (i.e., nodes) with over 30 million connections (i.e., edges) between persons, families, places, and various multimedia items. The datasets in this research contained genealogical trees that the Anu Museum holds consent and rights to publish online, complying with the European general data protection regulation[viii] (GDPR), and under the Israeli privacy regulation[ix]. Furthermore, the records of living individuals were removed from the datasets as much as possible. All personal data and any data that can be used to identify a person in this paper, including in the tables and figures, have been modified to protect the privacy of these individuals.

Based on the filtered GEDCOM files from the above corpus, and after removing some files with parsing or encoding errors, six datasets were generated for each dataset design described above. All datasets were split into training (60% - 941,809 questions), test (20% - 220,889 questions), and evaluation (20% - 220,888 questions) sets. It is worth mentioning that there was a difference in the efficiency of the dataset generation (i.e., the time that it takes to generate the dataset) between different dataset designs (e.g., GenAgg$_{6NF}$ requires more JOIN operations compared to GenAgg$_{agg}$); however, since the time of the dataset generation process is a small portion (less than 5%) of the time of the training process, it does not have a significant influence on the overall training efficiency.

### 4.2. Validation dataset creation

Due to the fact that the paraphrased questions in the dataset were generated automatically, it may contain errors. Therefore, a manually crafted dataset was created for the models' validation. To this end, a crowdsourcing campaign using Amazon Mechanical Turk was launched. A dedicated website was built that displayed randomly selected questions from the dataset along with the paraphrased versions of these questions. Each crowd worker evaluated up to ten items (i.e., pairs of original question vs. paraphrased question), and, for each item, answered the following questions:

1. Is the paraphrased question grammatically correct?
2. Does the paraphrased question preserve the meaning of the original question?

Before performing the task, each crowd worker got a "test task", as shown in Figure 12, to ensure they understood what needed to be done. If the crowd workers failed the test, they could not continue to perform the actual task. Each item was evaluated by three independent crowd workers, and a paraphrased question was considered correct / meaning preserving, only if at least two of the three crowd workers marked it as such. A total number of 667 crowd workers, who passed the test, were recruited and judged a random sample of 896 pairs of questions (1,792 questions in total) from the evaluation dataset. All participants signed a letter of consent before participating in the experiment. The IRB of the Faculty of Humanities at Bar-Ilan University has approved this study. Figure 12 presents the website's user interface that was built for this study.

Figure 12: Crowdsourcing website for building a validation dataset.

### 4.3. Fine-tuning the GLOBE QA model

The GLOBE numerical aggregative QA DNN model was first trained on the SQA [36] and WikiSQL datasets [87], replicating the TaPas training process. Then, the GLOBE QA model was fine-tuned using the generated training datasets, referred to as GenAgg. Each table in the datasets was lowercased and tokenized using WordPiece [83]. To evaluate the effect of the dataset design on the model's accuracy, the table selection DNN models and the numerical aggregative QA DNN models were trained on 1,383,586 questions for each of the six datasets. All the models were trained with the hyperparameters (similar to the TaPas hyperparameters) shown in Tables 5 and 6.

Table 5: Training hyperparameters for table selection DNN based on SBERT.

| Hyperparameters | Value |
| --- | --- |
| Max sequence tokens | 512 |
| Batch size | 18 |
| Training Steps | 25,000 |
| Loss | cosine similarity |
| Epocs | 10 |

Table 6: Training hyperparameters of GLOBE QA DNN.

| Hyperparameters | Value |
| --- | --- |
| Max sequence tokens | 512 |
| Batch size | 32 |
| Training Steps | 25,000 |
| Learning rate | 5e-5 |
| Epocs | 8 |

### 4.4. Accuracy metrics

To assess the models' accuracy, the following standard measures were employed in this study:

1. $Answerable_{acc}$ is the average accuracy of the model on questions that the model is able to answer (i.e., the tables required to answer these questions fit the model's input size). Accuracy is a common metric for QA systems [64].
2. $Total_{acc}$ is the total accuracy (for both answerable and unanswerable questions) calculated as follows:

$$Total_{acc} = Answerable_{acc} * (1 - \frac{Answerable_{quiestions}}{Total_{questions}})$$

where $Total_{questions}$ is the overall number of questions in the dataset.

In addition, for each of the above measures, we use both exact and soft evaluation approaches from previous research [32]. In particular, for a given question $q$, the answer $\hat{y}$ returned by the model, and the corresponding correct answer $y$, the exact and soft accuracy scores are computed as follows:

$$Exact_{acc}(\hat{y}, y) = \begin{cases} 1 & , \quad if\ \hat{y} = y \\ 0 & , \quad if\ \hat{y} \neq y \end{cases}$$

$$Soft_{acc}(\hat{y}, y) = \begin{cases} 1 & , \quad if\ \hat{y} = y \\ 0 & , \quad if\ \hat{y}\ is\ not\ a\ number \\ 1 - \dfrac{|\hat{y} - y|}{\max(\hat{y}, y)} & , \quad otherwise \end{cases}$$

## 5. EXPERIMENTAL RESULTS

To evaluate the proposed methodology, we computed the various models' accuracy on the evaluation dataset (i.e., 20% of the dataset that was not part of the training dataset). As a baseline for the comparative model evaluation, TaPas was trained on the GenAgg$_{1t}$ dataset (since it can only handle a single table per question-answer). As expected, it yielded a very low total accuracy (16-21%). As can be seen in Table 7, rich tables with predefined aggregations, as in GenAgg$_{event}$, improve the model's accuracy since they include the aggregations that otherwise need to be inferred by the model. However, joining data also leads to a high percentage of unanswerable questions (as for GenAgg$_{agg}$ and GenAgg$_{event}$), due to the fact that these large tables are often larger than the model's input size limit. When answering multi-table questions, the 6NF design creates simple and focused tables based on the given question, which dramatically increases the model's accuracy. Moreover, this design also reduces the table size and allows the model to answer almost any question. Overall, the 6NF design achieved the highest total accuracy (58-87%), which is 3-4 times higher than the baseline and other models. Furthermore, the dramatic increase in accuracy for all the models when using soft evaluation shows that when the GLOBE QA models are wrong, their predictions are very close to the correct answer.

Table 7: GLOBE models weak supervision accuracy

| QA model | Table selection model | Dataset | Exact accuracy of answerable questions | Soft accuracy of answerable questions | Unanswerable questions | Total exact accuracy | Total soft accuracy |
|---|---|---|---|---|---|---|---|
| TaPas$_{1t}$ | N\A | GenAgg$_{1t}$ | 23.49% | 30.57% | 31% | 16.21% | 21.09% |
| GLOBE$_{1t}$ | N\A | GenAgg$_{1t}$ | 24.95% | 32.47% | 20% | 19.96% | 25.97% |
| GLOBE$_{raw}$ | SBERT$_{raw}$ | GenAgg$_{raw}$ | 24.48% | 30.76% | 51.6% | 11.85% | 14.89% |
| GLOBE$_{rel}$ | SBERT$_{rel}$ | GenAgg$_{rel}$ | 25.67% | 32.67% | 48% | 13.35% | 16.98% |
| GLOBE$_{agg}$ | SBERT$_{agg}$ | GenAgg$_{agg}$ | 47.52% | 77.72% | 78.7% | 10.12% | 16.55% |
| GLOBE$_{event}$ | SBERT$_{event}$ | GenAgg$_{event}$ | 44.47% | 64.41% | 63% | 16.45% | 23.83% |
| GLOBE$_{6NF}$ | SBERT$_{6NF}$ | GenAgg$_{6NF}$ | 58.75% | 87.30% | 0.3% | **58.57%** | **87.04%** |

It is worth mentioning that the accuracy of the models depends on both the answerable questions' rate (i.e., questions that the model can answer with respect to its input size limit) and the accuracy of the answers.

To further improve the model's performance, we investigated the possible factors that could affect prediction accuracy. Thus, the quality of the question dataset may have a crucial effect on the model's results. The question dataset was automatically generated from the data tables based on predefined patterns and a paraphrasing DNN model. As shown in Table 8, the crowd workers' evaluation of the paraphrased questions in the dataset indicated that only 46.3% of them were grammatically correct and preserved the meaning of the original question. While the model should overcome some grammatical errors, it is harder to handle a change in the meaning of the question (e.g., when a condition is removed or modified).

Table 8: Crowd worker's evaluation results on the paraphrased question generation.

| Evaluation value | Paraphrased question is grammatically correct | Paraphrased question preserves the meaning of the original question | Paraphrased question is both grammatically correct and meaning preserving |
|---|---|---|---|
| Yes | 635 | 604 | 415 |
| No (otherwise) | 261 | 292 | 481 |
| **% correct** | **70.8%** | **67.4%** | **46.3%** |

To eliminate the impact of incorrect questions, the best GLOBE QA model (i.e., GLOBE$_{6NF}$) was validated only on questions marked as correct by the crowd workers. As shown in Table 9, there is a substantial increase (of over 20%) in the total exact accuracy when using high-quality questions. However, the small effect on the soft accuracy suggests that low-quality data divert the model prediction "just a bit" compared to the prediction on high-quality data. Moreover, the fact that there is only a small difference between the questions with correct grammar and questions with grammatical errors demonstrates the model's ability to overcome most grammatical errors. The validation accuracy displayed in Table 9 is the expected accuracy of the model in real-world applications with questions written by humans.

Table 9: GLOBE$_{6NF}$ validation accuracy.

| Dataset | Questions | Total exact accuracy | Total soft accuracy |
|---|---|---|---|
| GenAgg$_{6NF}$ evaluation | 220,889 | 58.57% | 87.04% |
| Validation dataset - preserved meaning | 1,208 | 78.84% | 87.52% |
| Validation dataset - preserved meaning and correct grammar | 830 | **80.15%** | **87.52%** |

Another possible influence factor on the model's accuracy could be the selected table quality. We assessed the accuracy of the table selection model using the evaluation dataset. As shown in Table 10, the simpler the dataset (i.e., fewer tables), the better the table selection model accuracy. Interestingly, one exception is the 6NF database, which contains a large number of tables.

Table 10: Table selection model accuracy.

| Model | Dataset | Total number of tables | Accuracy |
| --- | --- | --- | --- |
| $SBERT_{raw}$ | $GenAgg_{raw}$ | 2 | 63.93% |
| $SBERT_{rel}$ | $GenAgg_{rel}$ | 4 | 63.47% |
| $SBERT_{agg}$ | $GenAgg_{agg}$ | 4 | 61.26% |
| $SBERT_{event}$ | $GenAgg_{event}$ | 5 | 60.26% |
| $SBERT_{6NF}$ | $GenAgg_{6NF}$ | 77 | **87.5%** |

It is worth mentioning that in many cases, even if the table selection model was wrong, the QA model was able to predict the exact correct answer, since for many questions (in some datasets), there is more than one table that contains the answer. For instance, for the question "How many children do London residents have on average?", the answer can be calculated by counting the children of the fathers/mothers or by counting the fathers/mothers of the children. Moreover, there was no difference in accuracy when eliminating the DNN-based table selection errors by inputting only the correct tables into the model.

The aggregative operation may also be a possible influence factor on the model's accuracy. As can be seen in Table 11, there was a difference between different mathematical operations. While the exact accuracy is similar, there is a considerable difference in the soft accuracy. Operations with high soft accuracy (such as Count) show that the predictions were close to the correct answers even when the model was wrong. However, no single operation shows a significantly better (or worse) accuracy than the other operations.

Table 11: $GLOBE_{6NF}$ validation accuracy with different mathematical operations.

| Aggregative operation | Total exact accuracy | Total soft accuracy |
| --- | --- | --- |
| Count | 78.69% | 90.70% |
| Average | 79.43% | 81.89% |
| Minimum | 82.08% | 85.74% |
| Maximum | 78.53% | 83.21% |
| Portion | 81.76% | 88.27% |

Table 12 illustrates the accuracy of the top GLOBE models (i.e., GLOBE$_{6NF}$, GLOBE$_{event}$, GLOBE$_{agg}$). The table presents anecdotal examples of questions answered by the models.

Table 12: Answer prediction by the GLOBE models.

| Question | GLOBE$_{6NF}$ | GLOBE$_{event}$ | GLOBE$_{agg}$ | Correct answer |
|---|---|---|---|---|
| What is the average age of people born in Argentina? | 65.0 | unanswerable | unanswerable | 65.0 |
| What is the average age of people with the first name Sara and age of 45? | 45 | 45 | 45 | 60.4 |
| How many females first name is Shira? | 1 | 0.96 | 1 | 1 |
| What is the portion of females born in POLAND? | 0.1428 | 0.13 | 0 | 0.1410 |
| The maximum age of a person in Germany? | 105 | unanswerable | unanswerable | 105 |
| What is the portion of women with the last name Gershon and the age on the first child of 24? | 0.02 | 24 | 24 | 0.006 |
| What is the portion of people with the last name Gershon? | 0.15 | 0.142 | 0.142 | 0.1493 |
| How many people were born on average every year between 1850 to 1880? | 11.2 | 14.1 | unanswerable | 14.1 |
| How many people's birthplace is Kurdistan? | 3 | 3 | unanswerable | 3 |

While the best model (i.e., GLOBE$_{6NF}$) has the highest overall accuracy, in some cases, other models were able to provide correct answers to the questions for which the GLOBE$_{6NF}$ failed. However, no clear pattern was found for the questions that the GLOBE$_{6NF}$ failed to answer correctly (while other models succeeded). A deeper error analysis shows that, in some cases, when the question contains one number, and there is a cell in the table with that number, the models predict this cell as the only relevant one; this may suggest that the models need further training. When the size (i.e., the sequence length) of the tables required to answer the question is larger than the model's input size, the question is marked as unanswerable. Another type of errors occurs when the models fail to predict the aggregation function, thus yielding results that are not in the expected range (e.g., numbers larger than one for portion questions or lower than one for count questions).

Finally, to evaluate the added complexity of the GLOBE QA model (i.e., the effect of adding the portion aggregation function), both TaPas and GLOBE QA models were trained and evaluated on SQA (one of the open-domain datasets tested in [32]) that does not contain portion questions. When comparing the TaPas and GLOBE$_{1t}$ model's exact accuracy[x] on the SQA dataset (with weak supervision), only a slight insignificant difference was observed (78.1% vs. 77.8%).

## 6. DISCUSSION AND CONCLUSIONS

This paper outlines and implements an end-to-end multi-phase methodology for a novel challenging task of numerical aggregation QA in the genealogical domain. The presented methodology was evaluated using a large corpus of 1,847,200 different persons. The obtained results show that while on generic data (e.g., SQA), the GLOBE QA model seems to have no benefit, on geological data it outperformed the state-of-the-art model and achieved 80-87% accuracy. It also effectively implemented a new aggregation function highly popular for the genealogical research, portion, that was not supported in the previous research. This finding shows that the genealogy domain is distinctive in complexity, characteristics and requirements, and thus needs dedicated training methods, data modeling, and fine-tuned DNN models. The study also examined the dataset design's effect on the QA model's accuracy. The results show that the complexity of the genealogical domain requires a more complex pipeline that can split and reconstruct the data tables based on a question, where the most effective design is based on a 6NF approach. As expected, the results also indicate the importance of high-quality data and the negative effect of errors of automatic data augmentation (question paraphrasing) on the model's accuracy.

In summary, this study's contributions are: (1) the optimal table-based dataset design for the numerical aggregation QA for the genealogical domain (GenAgg$_{6NF}$); (2) an automated method for the training dataset generation for the genealogical domain; and (3) optimal fine-tuned table selection and numerical QA models for the genealogical domain (GLOBE QA).

The study may also have a substantial societal impact as genealogical centers, museums and various commercial organizations aim to allow users and experts without mathematical or programming training to investigate their large family tree databases. For example, imagine walking into a genealogical center and researching your own dynasty migration paths by asking questions like "How many people from my family were born in England but died in another county?", or a Holocaust researcher asking "What is the average marriage age of women in Germany between 1919-1939?" and then comparing the answer to the answer to the question "What is the average marriage age of women in Germany between 1945-1965?". To answer a variety of questions, such systems should incorporate factual and numerical aggregation QA based on the data stored within the GEDCOM files. Furthermore, practical and scientific implications of this study for the genealogical domain can be researching communities, migration, plagues, and marriage cultures all over the globe.

Future research may focus on (1) combining different methods for overcoming the sequence length limitation of the DNN models (e.g., [18, 19, 42, 85]), (2) improving the system's accuracy by developing a model selection or a multi-model method for various question types, (3) training other models (and performing hyperparameter optimization) on genealogical data to further optimize numerical QA models for the genealogical domain, (4) developing and comparing the proposed method to GNN model [90] by embedding nodes [91, 92] per relation degree [69], (5) adapting text-to-text models to present an explanation to the predicted answers [61, 52], (6) a deep error analysis of the models to identify possible improvements or using the models as a mix of experts (i.e., using different models for different questions or graphs), and (7) building the optimal user interface for the task of genealogical question-answering. Finally, in addition to the numerical aggregation QA task, the developed end-to-end methodology can also be applied to other downstream genealogical NLP (Natural Language Processing) tasks, including entity extraction, summarization, and classification.

## 7. ACKNOWLEDGMENTS

This work was partially supported by a grant from the Israel data science initiative (IDSI). The data of this work was granted for use by the Douglas E. Goldman Jewish Genealogy Center of Anu Museum with permission to publish statistical results.

---

i https://www.myheritage.co.il/about-myheritage/
ii https://support.ancestry.com/s/article/Searching-Public-Family-Trees
iii https://www.familysearch.org/en/about
iv First degree – parents and children; second degree – grandparents and grandchildren, etc.
v http://www.cidoc-crm.org/
vi Weakly supervised learning is a training method where the model receives only the final answer to the question, but no additional information, such as the aggregation function used to calculate the answer, the formal query or target cells leading to the answer.
vii https://dbs.anumuseum.org.il/skn/en/c6/e18493701
viii https://gdpr-info.eu/
ix https://www.gov.il/BlobFolder/legalinfo/data_security_regulation/en/PROTECTION%20OF%20PRIVACY%20REGULATIONS.pdf
x Since the goal of GLOBE and TaPas is not conversational (while SQA is a conversational dataset), we measured the exact accuracy on the first question in every question sequence (i.e., Q1).